\pdfoutput=1

\documentclass[11pt]{article}

\usepackage[]{acl}

\usepackage{times}
\usepackage{latexsym}
\usepackage{multirow}

\usepackage[T1]{fontenc}

\usepackage[utf8]{inputenc}

\usepackage{microtype}

\usepackage{inconsolata}

\usepackage{graphicx}

\usepackage{comment}
\usepackage{booktabs}
\usepackage{tabularx}
\usepackage{amsmath}

\usepackage{CJKutf8}
\usepackage[normalem]{ulem} 
\usepackage{adjustbox} 
\usepackage{booktabs} 

\usepackage{array}

\title{Instructions for *ACL Proceedings}
\title{Towards Comprehensive  Argument Analysis in Education: 

Dataset, Tasks, and Method}

\author{
    Yupei Ren\textsuperscript{$1$,$2$,$3$},
    Xinyi Zhou\textsuperscript{$4$},
    Ning Zhang\textsuperscript{$5$},
    Shangqing Zhao\textsuperscript{$3$},\\
    \textbf{Man Lan\textsuperscript{$1$,$2$,$3$}\thanks{\ \ Corresponding author.}},
      \textbf{Xiaopeng Bai\textsuperscript{$1$,$2$,$4$}}
    \\
    \textsuperscript{1}Lab of Artificial Intelligence for Education, East China Normal University \\
    \textsuperscript{2}Shanghai Institute of Artificial Intelligence for Education, East China Normal University\\
    \textsuperscript{3}School of Computer Science and Technology, East China Normal University \\
    \textsuperscript{4}Department of Chinese Language and Literature, East China Normal University \\
    \textsuperscript{5}College of Education, Zhejiang University \\
    \texttt{ypren@stu.ecnu.edu.cn, mlan@cs.ecnu.edu.cn} \\
}
\begin{document}
\maketitle
\begin{abstract}
Argument mining has garnered increasing attention over the years, with the recent advancement of Large Language Models (LLMs) further propelling this trend. However, current argument relations remain relatively simplistic and foundational, struggling to capture the full scope of argument information, particularly when it comes to representing complex argument structures in real-world scenarios. To address this limitation, we propose 14 fine-grained relation types from both vertical and horizontal dimensions, thereby capturing the intricate interplay between argument components for a thorough understanding of argument structure. On this basis, we conducted extensive experiments on three tasks: argument component detection, relation prediction, and automated essay grading. Additionally, we explored the impact of writing quality on argument component detection and relation prediction, as well as the connections between discourse relations and argumentative features. The findings highlight the importance of fine-grained argumentative annotations for argumentative writing quality assessment and encourage multi-dimensional argument analysis.
\end{abstract}

\section{Introduction}

\begin{figure}[t]
\centering
\includegraphics[width=0.95\columnwidth]{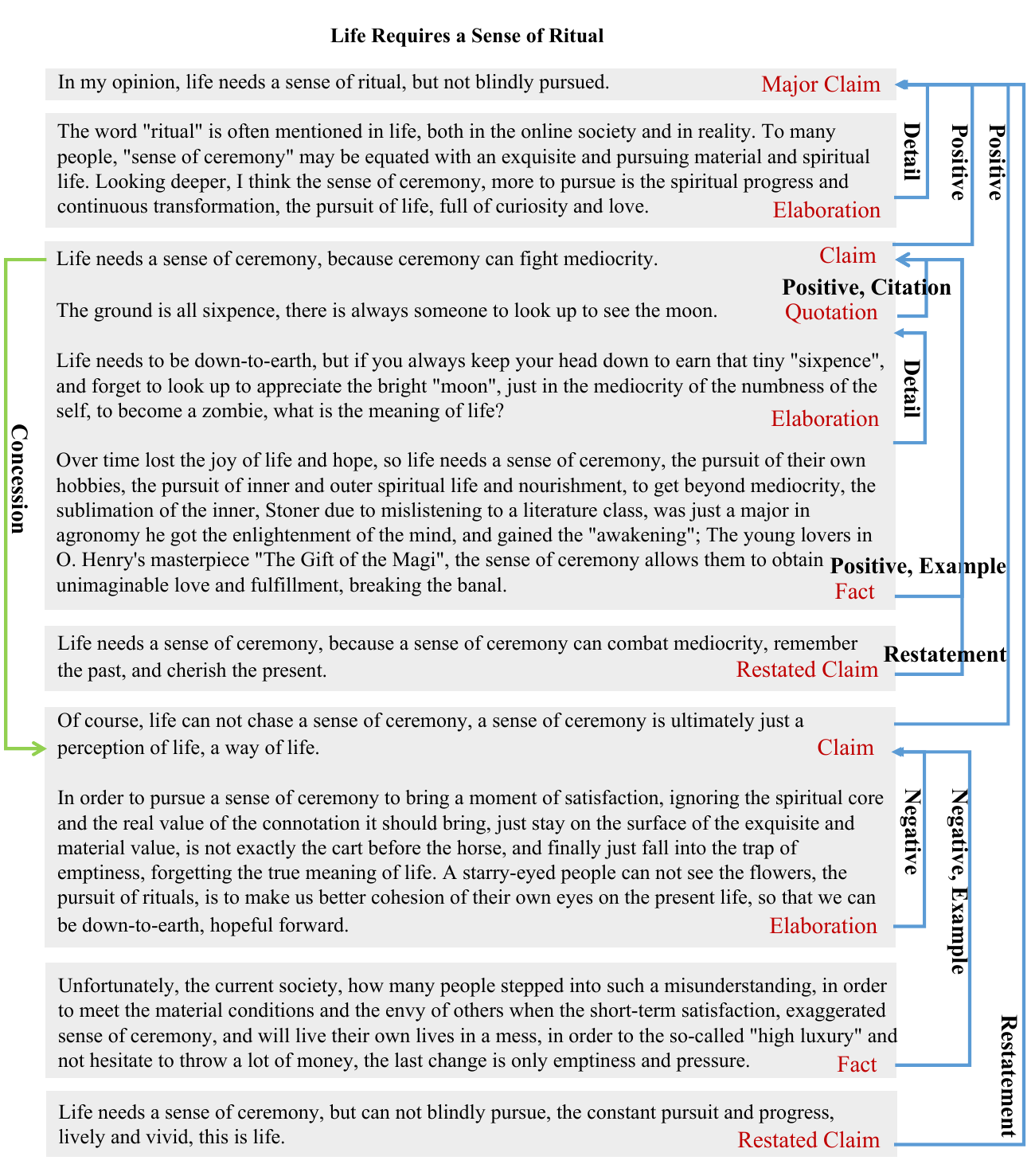} 
\caption{An Annotation Example (Excerpt). The red font indicates argument component types, the blue arrows on the right signify vertical argument relations, and the green arrow on the left represent horizontal logical relations. The content above the arrows corresponds to the respective relation types.}
\label{fig: Data sampling}
\end{figure}

Argument Mining (AM) aims to automatically extract structured argumentation information from unstructured texts, encompassing the analysis of argument units, comprehending their roles and interactions within a document, and ultimately forming a cohesive argumentation \cite{lippi2016argumentation}. The automatic identification of argument structure holds significant promise, providing valuable support for various downstream Natural Language Processing (NLP) tasks such as quality evaluation\cite{stahl2024school} and text generation \cite{lin2023argue, chen2023exploring}. 

Existing research on argument mining has proposed various argument annotation schemes and tasks, mainly focusing on two aspects: (a) \textit{Argument Component Detection}, and (b) \textit{Argument Relation Prediction}. Each element embodies a unique aspect of argument mining. Regarding argument component detection, \citet{guo-etal-2023-aqe} emphasize that a comprehensive understanding of argumentative texts requires knowledge of the viewpoints (i.e., claims) presented in the text, the validity of those viewpoints (i.e., supporting evidence), and the source of the evidence (i.e., evidence types). In terms of argument relation prediction, \citet{mochales2011argumentation} argue that the core of argument analysis lies in comprehending the content of the argument chains, analyzing linguistic structures, and determining the relations between argument units to reveal the argument structure of the text.

Despite significant progress in the field of argument mining, the current argument relations are still relatively simple and basic, making it difficult to capture complete argument information, especially to meet the representation needs of complex argument structures in real scenarios. For example, most argument mining studies \cite{cheng2022iam, schaller2024darius} only categorizes argument relations into \textit{support} and \textit{attack} based on stances, lacking the characterization of critical information such as argument strategies and patterns, which are essential for a thorough understanding and evaluation of the overall structure and quality of arguments. Furthermore, existing quality assessment of argumentative essays primarily focuses on scoring annotations for argumentative attributes such as strength, relevance \cite{wambsganss2022modeling}, content, and style \cite{schaller2024darius}. While these innovative annotation methods enhance the granularity of assessment, they overlook the intrinsic value of argument components and relations as key argumentative features, hindering the effective integration of argument component detection and relation prediction with quality assessment.

To address the shortcomings of existing research, we propose an innovative relation annotation scheme
to characterize the argument strategies and patterns within the \textbf{C}hinese \textbf{E}ssay \textbf{A}rgument \textbf{M}ining \textbf{C}orpus (\textbf{CEAMC}) \cite{ren2024ceamc}. As shown in Figure \ref{fig: Data sampling}, each argumentative essay undergoes meticulous annotation. We argue that these annotations address the key limitations in prior work: 
\textbf{First}, it overcomes the issue of simplified argument relations prevalent in previous studies. By deeply integrating argument relations with discourse relations, it introduces 14 fine-grained relation types from both vertical and horizontal dimensions, comprehensively depicting the complex interactions between argument components and providing a deeper understanding of argument structures. 
\textbf{Second}, it breaks away from the isolationist approach previous studies. With the integration of argument component, relations, and essay grading, it provides a more comprehensive understanding of argument analysis. 
\textbf{Lastly}, the detailed annotations adeptly capture the subtle nuances of physical real-world argumentative texts, providing a more reliable basis for argument evaluation and instruction.

Our contributions can be summarized as follows:
\begin{itemize}

\item We have revised and enhanced a comprehensive multi-task dataset for argument analysis, enhancing understanding of Chinese high school student argumentative essays.

\item We provide comprehensive benchmarks for each task, systematically evaluate the performance of existing methods, and offer reference points for future research.

\item Through insightful experiments, we illustrate the impact of writing quality on argument component detection and relation prediction, and explore the connections between discourse relations and argumentative features, encouraging multi-dimensional argument analysis.

\end{itemize}

\section{Related Work}

\subsection{Argument Mining}

Most argument mining studies \cite{fergadis2021argumentation, wambsganss2022modeling,jundi-etal-2023-node} have focused on identifying the basic argument components and relations, namely the three components of \textit{major claim}, \textit{claim} and \textit{premise}, as well as the two relations of \textit{support} and \textit{attack}.

Existing studies have delved into argument components, including refining the categories from the perspective of sentence functions \cite{song2021hierarchical, kennard2022disapere} and further categorizing them based on evidence attributes \cite{niculae2017argument, guo-etal-2023-aqe}. Recently, research on argumentative essays in German \cite{schaller2024darius, stahl2024school} and Chinese  \cite{ren2024ceamc} schools has advanced the study of argumentation education through multi-level granularity annotation.  While these efforts have facilitated an understanding of arguments, they lack a thorough exploration of argument relations.


Regarding argument relations, several studies have refined additional relations based on discourse relations, such as \textit{detail}, \textit{sequence} \cite{kirschner2015linking}, \textit{by-means}, \textit{info-required} and \textit{info-optional} \cite{accuosto2021argumentation}, which hold significant value in scientific literature. Similarly, \citet{jo2021classifying} adopted \textit{causal} and \textit{normative} relations to complement the relations in debates. Recently, \citet{liu2024antcritic} defined \textit{affiliation}, \textit{co-occurrence} and \textit{co-relevance} relations to characterize the argument structure within documents from online financial forums. While these efforts have enriched argument relations and advanced the understanding of argument structure, they still lack key argument information such as argument strategies and patterns. Furthermore, most of these findings are concentrated in out-of-education domains and mainly in English and German, limiting the further development and application of argument mining research.

\subsection{Discourse Relation Recognition}

Discourse Relation Recognition (DRR) aims at detecting semantic relations between text units, thereby modeling the logical structure of discourse. Existing research on discourse relations is mainly based on Rhetorical Structure Theory (RST) \cite{mann1988rhetorical} and Penn Discourse Treebank (PDTB) \cite{prasad2008penn}. On this basis, considering the nuances of Chinese discourse, \citet{wu2023multi} formulated a framework consisting of  four tiers and thirteen labels. This framework encompasses a wide range logical semantic types in Chinese discourse and promotes the development of discourse relation research.

It is noteworthy that discourse structure is closely associated with argument structure, and the research of discourse relations plays a critical role in guiding argument mining. \citet{cabrio2013discourse} and \citet{stab2017parsing} have both emphasized the significance of discourse relations in argument mining research and advocate for combining the two to foster insightful investigation.

\section{Corpus Construction}

In this section, we briefly describe the corpus we use, i.e. the CEAMC corpus \cite{ren2024ceamc}. In addition, we present our annotation scheme, the procedure, and results.

\subsection{Source Data}

The CEAMC corpus \cite{ren2024ceamc} comprises 226 argumentative essays from high school examination context, annotated with 4 coarse-grained and 10 fine-grained sentence-level argument components
(i.e., \textit{Assertion}: \textit{major claim}, \textit{claim} and \textit{restated claim}; \textit{Evidence}: \textit{fact}, \textit{anecdote}, \textit{quotation}, \textit{proverb}, and \textit{axiom}; \textit{Elaboration}; and \textit{Others}). 
Essays from authentic educational settings encapsulate rich argumentative information, offering a unique perspective for insightful exploration of argument strategies and structures. 
Based on this, we conduct extensive relation annotations. The source data also includes essay score information.

\subsection{Annotation Scheme}

In persuasive writing, different argument components and targets lead to various argument strategies. To provide a comprehensive representation and profound analysis of argumentative essays, we innovatively annotate the relations within argumentative essays from both vertical and horizontal dimensions, based on education practice and by integrating both argument and discourse relations.

\subsubsection{Vertical Dimension} 
The vertical dimension focus on the relations between different types of argument components, aiming to reveal the internal logic and reasoning chains of arguments. We have defined ten types of argument relations from three aspects to comprehensively characterize the collaborative interactions between argument components.

\paragraph{Stance-Based Argument Relations} Most argument mining research categorizes argument relations into support and attack based on stance, but occurrences of attack relations are quite rare \cite{stab-gurevych-2014-annotating, stab2017parsing, stahl2024school}. Additionally, no attack relations were observed in the analysis of online comments by \citet{park2018corpus}, nor in the research on business proposals by German university students conducted by \citet{wambsganss2022modeling}. Furthermore, \citet{song2021hierarchical} did not annotate the relations in the Chinese online argumentative essays and subtly implied that there was a support relation between the evidence and the claim. Taking into account the fact that when writing persuasive essays, students aim to argue for their major claims in the most persuasive manner, typically without overemphasizing attack relations between argument components, but more commonly from the opposite side to strengthen their reasoning. Based on these observations, we propose three stance-based argument relations, namely \textbf{Positive}, \textbf{Negative}, and \textbf{Comparative} argumentation, to understand students' argumentative essays.

\paragraph{Evidence-Based Argument Relations} Different types of evidence lead to different modes of argumentation. In conjunction with educational practice, evidence-based argument relations include \textbf{Example} and \textbf{Citation} argumentation.

\paragraph{Discourse-Based Argument Relations} 

To comprehensively depict the argumentation process of students, we integrate discourse analysis theory to further expand and refine the argument relations. Drawing on the framework of RST \cite{mann1988rhetorical}, we introduce three new categories of relations, namely \textbf{Background}, \textbf{Detail}, and \textbf{Restatement}, to enhance the understanding of argument structure. Based on the research by \citet{wu2023multi} on Chinese discourse relations, we further define \textbf{Hypothetical Argumentation}, which plays a significant role in Chinese argumentation. Moreover, considering the importance of metaphoric rhetoric in argumentative activities \cite{pilgram2021strategic}, we introduce \textit{Metaphorical} Argumentation.

\subsubsection{Horizontal Dimension} 

The horizontal dimension focuses on the relations between argument components of the same type, aiming to analyze the interconnections between elements at the same level (such as the relations between claims and how they collectively support the major claim), thus facilitating a comprehensive understanding of argument structure. Grounded in Chinese argumentation teaching, we draw on the research by \citet{wu2023multi} and utilize four discourse relations, namely \textbf{Coherence}, \textbf{Progression}, \textbf{Contrast}, and \textbf{Concession}, to annotate the logical transitions between arguments of the same type.

For a detailed overview of relation types and samples, please refer to Appendix \ref{Annotation Scheme}.

\subsection{Annotation Process}

Annotation team consists of three undergraduates, three postgraduates specializing in linguistics and education, and two experts with extensive experience in Chinese teaching. Before the formal annotation work, the team underwent a series of training sessions and pre-annotation exercises to better familiarize and master the task requirements. Building on this, we discussed their understanding of the guidelines and variations in annotations, making appropriate adjustments to the guide. Each essay was independently annotated by two annotators, with domain experts responsible for coordinating and resolving any disagreements between them.

It is noteworthy that, before undertaking the relation annotation task, we asked two experts to clearly demarcate the boundaries of argument units based on the results of sentence-level argument component annotations. This step was particularly crucial because we observed that in the Chinese context, there are often multiple consecutive sentences of the same type discussing the same content, which poses challenges to annotating relations between arguments. On this basis, we annotated the relations from both vertical and horizontal dimensions, encompassing 226 argumentative essays with a total of 4,837 relations.

\subsection{Inner Annotator Agreements}

We followed \citet{cheng2022iam} and \citet{liu2024antcritic}, using Cohen’s kappa to measure Inter-Annotator Agreement (IAA). A total of 3,458 argument units were derived from 4,726 sentences, with an IAA score of 0.95 for the annotation of argument unit boundaries, which indicates a high degree of consistency, providing reliable outcomes for subsequent relation annotation. Based on this, a total of 4,837 relations were annotated with an IAA score of 0.68, which is a reasonable and relatively high agreement considering the diversity and complexity of relation annotations.

\subsection{Data Statistics}

The final corpus consists of 226 Chinese argumentative essays, comprising 3,458 argument units, 3,923 argument pairs, and 4,837 relations (multiple relations may exist between each argument pair). As shown in Table \ref{table:relation distribution}, there are significant differences in the distribution of various argument relations and discourse relations, indicating that students exhibit diversity and complexity in mastering argument structures and relations in argumentative writing.

\begin{table}[thb]
\centering
\resizebox{\linewidth}{!}{
\begin{tabular}{l|l|l|r|r}
\toprule
\textbf{Dimension} & \textbf{Aspect} & \textbf{Label} & \textbf{Counts} & \textbf{\% of Total}  \\
\midrule
Vertical   & Stance-Based          & Positive           & 1,599    & 33.04\%       \\
    (4,102)                &                       & Negative           & 396      & 8.19\%        \\
                   &                       & Comparative        & 27       & 0.56\%        \\
\cmidrule{2-5}
                   & Evidence-Based        & Example            & 661      & 13.67\%       \\
                   &                       & Citation           & 216      & 4.47\%        \\
\cmidrule{2-5}
                   & Discourse-Based       & Metaphorical       & 31       & 0.64\%        \\
                   &                       & Hypothetical       & 6        & 0.12\%        \\
                   &                       & Restatement        & 203      & 4.20\%        \\
                   &                       & Detail             & 698      & 14.43\%       \\
                   &                       & Background         & 265      & 5.48\%        \\
\midrule
Horizontal    & -                     & Coherence          & 277      & 5.73\%        \\
 (735)                  &                      & Progression        & 305      & 6.31\%        \\
                   &                      & Contrast           & 46       & 0.95\%       \\
                   &                      & Concession         & 107      & 2.21\%       \\
\midrule
Total              & -                     & -              & 4,837     & 100.00\%  \\
\bottomrule
\end{tabular}
}
\caption{Distribution of relations. 
}
\label{table:relation distribution}
\end{table}

\section{Experiments}

\subsection{Tasks} \label{task}

Our annotated dataset serves as the foundation for three core tasks, each delving into distinct facets of argument analysis:

\begin{table}[htbp]
\centering
\footnotesize
\resizebox{\linewidth}{!}{
\begin{tabular}{lcccc}
\toprule
& \textbf{Train} & \textbf{Dev.} & \textbf{Test} & \textbf{Total} \\
\midrule
\multicolumn{5}{c}{\textit{Argument Component Detection }} \\
\midrule
\# Sentences & 3,805 & 451 & 470 & 4,726 \\
\# Arguments & 2,767 & 346 & 345 & 3,458 \\

\midrule
\multicolumn{5}{c}{\textit{Relation Prediction}} \\
\midrule
\# Positives  & 3,133 & 398 & 392 & 3,923 \\
 \hspace*{1em} \# Relations & 3,866 & 484 & 487 & 4,837 \\
\# Negatives & 16,410 & 1,992 & 2,039 & 20,441 \\

\midrule
\multicolumn{5}{c}{\textit{Automated Essay Grading}} \\
\midrule
\# Essays & 180 & 23 & 23 & 226 \\
\bottomrule
\end{tabular}}
\caption{Data split statistics for benchmark testing.}
\label{table:Data split statistics for benchmark testing}
\end{table}

\textbf{Argument Component Detection}. This task aims to detect and classify all potential argument components. We formulate it as a sentence-level classification task, utilizing IOB tagging to represent structural span information.

\textbf{Relation Prediction}. This task aims to detect and classify all relations between argument components. We frame it as argument-pair classification task: given a pair of argument components, predict the types of relations between them, noting the multi-label nature due to multiple relation types.

\textbf{Automated Essay Grading}. This task aims to evaluate the overall quality of students' argumentative essays. We frame it as a four-classification task, with detailed writing proficiency levels provided in Appendix \ref{Details of Essay Grading}.

To address above tasks, we split our data as summarized in Table \ref{table:Data split statistics for benchmark testing}. Across all tasks, a total of 226 labeled argumentative essays are split in an approximate 8:1:1 ratio. 
It should be noted that in the second task of relation prediction, our data statistics indicates that setting an argument distance of 15 covers almost 99\% of the positive argument pairs. Therefore, we construct negative samples based on a forward-backward distance of 15 and argument component types, that is, argument pairs with no existing relations.

\subsection{Baselines and Metrics}

We experiment on two well-established pretrained language models (PLMs): \textit{BERT} \cite{kenton2019bert} and \textit{RoBERTa} \cite{liu2019roberta}. Given the recent unparalleled achievements of LLMs in various NLP tasks, we also employ the LoRA technique \cite{hu2021lora} to conduct supervised fine-tuning (SFT) on three open-source Chinese LLMs, \textit{Qwen} \cite{yang2024qwen2}, \textit{DeepSeek} \cite{guo2025deepseek}, and \textit{ChatGLM} \cite{glm2024chatglm}, to evaluate their performance on each of our argument analysis tasks. Additionally, we assess the performance of OpenAI's ChatGPT\footnote{\label{openai_url}\url{https://openai.com/blog/chatgpt}}, specifically \textit{GPT-4-turbo}, under zero-shot and few-shot prompting conditions, to serve as a reference.

\textbf{Argument Component Detection:} We fine-tune various PLMs and LLMs on the training dataset, leveraging their powerful language modeling capabilities. We evaluate the performance of models using Precision ($P$), Recall ($R$), $F_{1}$-score ($F_{1}$). More precisely, the true positive for calculation is defined as the number of predicted argument components that exactly match a gold standard argument component, i.e., their boundaries and category labels are identical.

\textbf{Relation Prediction:} We evaluate the performance of models using Micro-$F_{1}$, Macro-$F_{1}$, and Pos.-$F_{1}$. Precisely, Pos.-$F_{1}$ aims to measure the models' ability to identify positive samples with relations, focusing solely on whether a relation exists between arguments, without distinguishing the specific types of relations. It is noteworthy that the argument pairs labeled with no-relation far exceeds other types of relations (as shown in Table \ref{table:Data split statistics for benchmark testing}). To overcome this challenge, we adopt negative sampling techniques \cite{mikolov2013distributed}. 
During the training process, we randomly select a certain amount of unrelated arguments for each argument as negative samples. These negative samples, along with all other arguments, form a new training dataset.

\textbf{Automated Essay Grading:} In addition to using the original essay as input, we incorporate argument components and relations to explore the impact of fine-grained argumentative information on essay grading. We evaluate model performance using $P$, $R$, $F_{1}$, Accuracy ($Acc$), and Quadratic weighted Kappa (QWK) \cite{vanbelle2016new}.

\subsection{Implementation Details}

For PLMs, we implement BERT-Base-Chinese and Chinese-RoBERTa-wwm-ext, using an AdamW optimizer with a learning rate of 2e$^{-5}$ to update the model parameters, and set the batch size to 8. For the open-source LLMs, we use Qwen2-7B-Base, DeepSeek-R1-Distill-Qwen-7B, and ChatGLM-4-9b-Base, employing LoRA throughout all training sessions with a LoRA rank of 8 and a dropout rate of 0.1. Training configurations include the learning rate of 5e$^{-5}$ and the batch size of 2. 
All our experiments are conducted on a single NVIDIA RTX 3090 GPU. 
All other parameters are initialized with the default values in PyTorch Lightning\footnote{https://github.com/Lightning-AI/lightning}, and our models are entirely implemented by Transformers\footnote{https://github.com/huggingface/transformers}.

\begin{table}[t]
\centering
\footnotesize
\begin{tabular}{lccc}
\toprule
    \textbf{Model}  & \textbf{$P$(\%)} & \textbf{$R$(\%)} & \textbf{$F_{1}$ (\%)} \\ \midrule
   BERT   & 40.05     &  47.83    & 43.59  \\
    RoBERTa   & 46.34     & 51.30     & 48.69 \\ \midrule
    Qwen$_{ft}$  &57.40   & 56.23   & 56.81  \\
    DeepSeek$_{ft}$     &  53.23   &  50.14    &    51.64     \\
    ChatGLM$_{ft}$ & \textbf{58.17}     & \textbf{58.84}     & \textbf{58.50} \\ \midrule 
    GPT-4$_{0-shot}$ & 29.50     & 34.20     & 31.68        \\
GPT-4$_{1-shot}$ & 27.01     & 27.44     & 27.19       \\ 
GPT-4$_{3-shot}$ & 32.66     & 33.04     & 32.85       \\ 

\bottomrule
\end{tabular}
\caption{Results for Argument Component Detection. }
\label{table: task1_results}
\end{table}

\subsection{Results and Analysis}

\subsubsection{Argument Component Detection}

Table \ref{table: task1_results} showcases the performance of various models on the \textit{Argument Component Detection} task. In the SFT setting, LLMs outperform PLMs on all metrics, demonstrating the exceptional ability of LLMs in identifying and predicting argument components. This superiority is attributed to the extensive knowledge and learning capabilities of LLMs, confirming the scaling laws that larger models tend to yield better performance \cite{kaplan2020scaling}. Furthermore, the larger parameter-sized ChatGLM (9b) surpasses Qwen (7B) and DeepSeek (7B) in achieving the best performance, which further corroborates this principle.

In contrast, while GPT-4 performs commendably under 0-shot and few-shot conditions, it significantly lags behind the models under SFT, highlighting the advantages of SFT and the importance of data annotation. Moreover, adding prompt examples does not significantly enhance GPT-4's deep understanding of the task, with its performance visibly diminishing in the 1-shot setting while showing only a slight improvement in the 3-shot setting. This seems to confirm the sensitivity and instability of LLMs to prompt samples, suggesting that procuring high-quality samples to improve the performance of LLMs warrants further investigation.

\subsubsection{Relation Prediction} 

Table \ref{table: task_2 results} displays the performance of various models on the task of \textit{Relation Prediction}, where 1 negative sample is randomly selected for each argument. It is clear that LLMs' performance in Micro-$F_{1}$ is comparable to that of PLMs, yet markedly superior in Macro-$F_{1}$ and Pos.-$F_{1}$. This demonstrates LLMs' proficiency in relation prediction, particularly in identifying positive samples with existing relations and handling imbalanced data. 
It is noteworthy that GPT-4 scores very low in both Micro-$F_{1}$ and Macro-$F_{1}$ under all settings, indicating significant challenges in relation prediction. Analysis of outputs reveals a tendency to mistakenly classify negative samples as positives, yet in reality, negative samples without relations far outnumber those with relations. This may stem from GPT-4’s extensive understanding of relations, which exceeds the scope of our defined argument pair relations, leading to its classification of most data as positives. This further emphasizes the importance of domain-specific fine-tuning of LLMs with annotated data.

\begin{figure}[t]
\centering
\includegraphics[width=0.95\columnwidth]{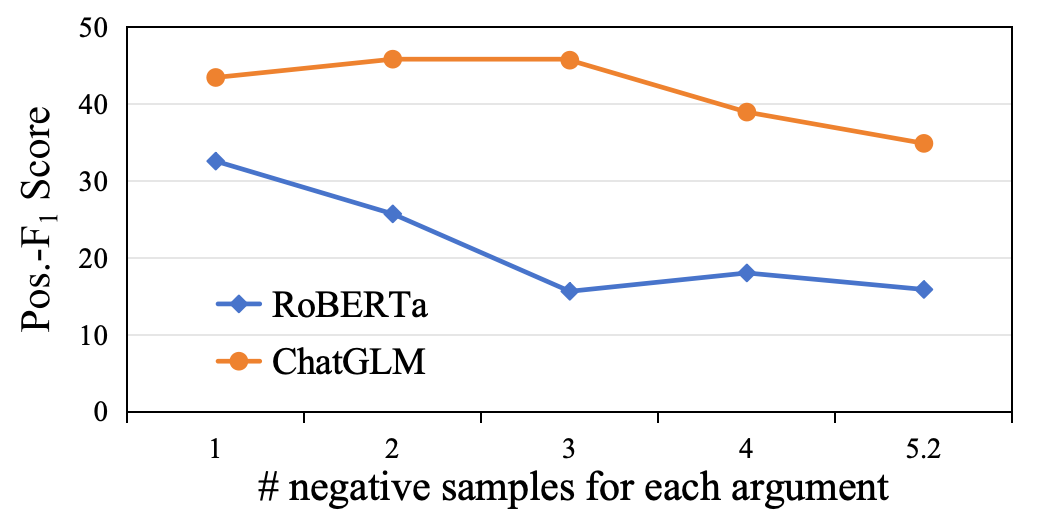} 
\caption{Effect of negative sampling for \textit{Relation Prediction} task with RoBERTa and ChatGLM models.}
\label{fig: neg sampling}
\end{figure}

As depicted in Figure \ref{fig: neg sampling}, we also compare the performance of the RoBERTa and ChatGLM models with varying numbers of negative samples per argument. Interestingly, ChatGLM peaks in performance with 3 negative samples, while RoBERTa shows the worst performance at this sampling size, reflecting significant differences between LLM and PLM. This may be due to the LLMs' extensive pre-training data and larger parameter size, which allow for better generalization and learning from more diverse data. Conversely, RoBERTa, as a smaller PLM, potentially suffers from an insufficient capacity for abstraction when processing larger numbers of negative samples, particularly in the context of complex relation judgments, thus exhibiting a marked decrease in performance as the number of negative samples rises. Notably, both models exhibit poor performance when the proportion of negative samples is excessively high. These insights highlight the importance of considering model size and learning capabilities when designing sampling strategies.

\begin{table}[t]
\centering
\footnotesize
\resizebox{\linewidth}{!}{
\begin{tabular}{lccc}
\toprule
   \textbf{Model} & \textbf{Micro-$F_{1}$} & \textbf{Macro-$F_{1}$} & \textbf{Pos.-$F_{1}$} \\ \midrule
    BERT  & \textbf{67.67}     &  16.45    & 28.81  \\
    RoBERTa   & 65.45     & 19.29     & 32.58  \\\midrule
    Qwen$_{ft}$     & 64.29     & 25.20    & 41.78        \\ 
    DeepSeek$_{ft}$     &  64.95   &  23.40    &    38.63      \\ 
    ChatGLM$_{ft}$  &66.78    & \textbf{32.68}    & \textbf{43.01} \\  \midrule

    GPT-4$_{0-shot}$ & 2.64     & 4.65    & 27.82          \\
    GPT-4$_{1-shot}$ & 10.61     & 5.72     & 28.14         \\ 
    GPT-4$_{3-shot}$ & 4.97    & 4.89     & 27.94        \\ 

\bottomrule
\end{tabular}}
\caption{Results for Relation Prediction task.}
\label{table: task_2 results}
\end{table}

\subsubsection{Automated Essay Grading} \label{sec: reselt of aes}

Table \ref{table: AES result} showcases the performance of various models on the Automated Essay Grading task. It is noteworthy that the models achieved promising results in this four-classification task, demonstrating the strengths of current methods in assessing students’ overall writing proficiency. Under the SFT setting with essays as input, PLMs generally outperformed LLMs. Analysis of the prediction results indicates that LLMs tend to give lower writing ratings. This may be due to the high proportion of high-quality argumentative essays in the training data, while high school students’ argumentative writing skills are still developing and generally below the adult level, leading to stricter evaluations by LLMs. Additionally, the scarcity of data somewhat limits the performance of LLMs in domain-specific tasks.

When fine-grained argument components and relations information were incorporated into the input, Longformer achieved the best QWK results, while QWen and DeepSeek also exhibited significant improvements, except for a decline in ChatGLM’s performance. This suggests that fine-grained annotation can enhance model performance in writing evaluation to a certain extent. However, further exploration is needed to effectively leverage fine-grained argumentative information to unlock the potential of LLMs in automated essay grading.

Notably, GPT-4 performed exceptionally well under 0-shot conditions, even surpassing the fine-tuned BERT model. However, its performance decreased significantly in 1-shot and 3-shot settings. This phenomenon not only highlights LLMs’ sensitivity to input examples but also underscores their misalignment with real-world educational scenarios in high school argumentative writing assessment. These findings suggest that specific task design requires a comprehensive consideration of domain-specific characteristics, task complexity, data scale, and model capabilities.

\begin{table}[]
\tabcolsep=0.4cm
\centering
\resizebox{\linewidth}{!}{
\begin{tabular}{@{}lccccc@{}}
\toprule
\textbf{Model}              & \textbf{$P$ (\%)} & \textbf{$R$ (\%)} & \textbf{$F_{1}$ (\%)} & \textbf{$Acc$ (\%)} & \textbf{QWK} \\ \midrule
BERT              & 63.89    &  61.90    & 62.22         & 69.57   & 0.5756    \\
RoBERTa           & 84.13   & \textbf{84.13}  & \textbf{84.13}   & \textbf{82.61}    & 0.8296  \\ 
Longformer        & 84.13  & \textbf{84.13}  & \textbf{84.13}   & \textbf{82.61} & 0.8296 \\ \midrule
Qwen$_{ft}$     & 78.82     & 58.73     & 62.81     & 73.91   & 0.5430 \\
DeepSeek$_{ft}$     & 47.44     & 56.35     & 50.46       & 73.91   & 0.6584 \\ 
ChatGLM$_{ft}$     & 75.00     & 73.02     & 73.33        & 73.91  & 0.7454 \\ 
 \midrule
Longformer$^{\dagger}$  & \textbf{85.42}  & 80.95 & 82.22 & \textbf{82.61} & \textbf{0.8303} \\
Qwen$^{\dagger}_{ft}$  &  76.67    & 67.46   & 70.92    &73.91  &  0.7218\\
DeepSeek$^{\dagger}_{ft}$  & 74.56     & 61.11     & 63.28         &73.91    & 0.7240 \\
ChatGLM$^{\dagger}_{ft}$  & 67.97    & 67.46     & 66.55        & 69.57   & 0.7035 \\ \midrule
GPT-4$_{0\text{-}shot}$ &64.44     & 74.60     & 65.97        & 65.22  & 0.5952   \\
GPT-4$_{1\text{-}shot}$ & 46.19     & 51.59     & 42.91        & 39.13   & 0.2014    \\ 
GPT-4$_{3\text{-}shot}$ & 35.07     & 33.33     & 33.07        & 43.48   & 0.2515    \\\bottomrule
\end{tabular}
}
\caption{Results for Automated Essay Grading. $^{\dagger}$ denotes the incorporation of annotation information.}
\label{table: AES result}
\end{table}

\section{Discussion} \label{Discussion}

This study investigates the significance of fine-grained annotation for the understanding of argumentation. Specifically, we explore the impact of writing quality on Argument Component Detection (Task 1) and Relation Prediction (Task 2). Additionally, we employ the learning analytics method ENA (Epistemic Network Analysis) \citep{shaffer2016tutorial} to compare the differences in horizontal discourse relations and vertical argument relation between high- and low-quality essays, aiming to visualize and provide interpretable insights into discourse relations and argumentative features (see Appendix \ref{ENA} for relevant concept). The grouping criteria are described in Appendix \ref{Details of Essay Grading}.

\subsection{The Impact of Writing Quality on Argument Component Detection and Relation Prediction} \label{dis 1}

According to Table \ref{table:Impact of Writing Performance}, while the performances varies between different writing levels, certain patterns are evident. \textbf{In Task 1}, models perform better on high-quality essays. Specifically, compared to low-quality essays, RoBERTa demonstrates significantly superior performance on high-quality essays, while ChatGLM and GPT-4 also show slight advantages. This indicates that high-quality essays typically possess clearer argument structures, which enhances the models’ ability to identify argument components. It also suggests that LLMs, leveraging their extensive knowledge, can partially mitigate the impact of writing quality differences on argument component detection. In contrast, \textbf{Task 2} yields markedly different results. In most cases, the models perform significantly better in relation identification for low-quality essays compared to high-quality ones, suggesting that the more complex and diverse relationships and structures in high-quality argumentative writing pose greater challenges to the models' predictive capabilities.

These findings suggest that the impact of writing quality on model performance varies depending on task type and difficulty, underscoring the importance of considering writing proficiency in argument component detection and relation prediction tasks. This echoes the result in Section \ref{sec: reselt of aes} that fine-grained argument information can assist in predicting writing proficiency, collectively revealing the intricate relationships among writing quality, argument information, and model performance.

\begin{table}[htb]
    \centering
    \resizebox{\linewidth}{!}{
    \begin{tabular}{l|c|ccc|ccc}
        \toprule
        \multicolumn{1}{c|}{\multirow{2}{*}{\textbf{Model}}} & \multicolumn{1}{c|}{\multirow{2}{*}{\textbf{Grade}}} &\multicolumn{3}{c|}{\textbf{Task 1}} & \multicolumn{3}{c}{\textbf{Task 2}}  \\

         &  & \textbf{$P$(\%)} & \textbf{$R$(\%)} & \textbf{$F_{1}$ (\%)} & Micro-$F_{1}$ & Macro-$F_{1}$ & Pos.-$F_{1}$  \\
        \midrule
       
        RoBERTa & High & \textbf{49.71} & \textbf{54.84} &  \textbf{52.15} & 64.58 & 18.76 & 31.13 \\
        
         & Low & 43.60 & 48.42 & 45.89 & \textbf{66.18} & \textbf{19.21} &  \textbf{33.80} \\
        \midrule    
        ChatGLM &  High & \textbf{58.78} & 58.78 & \textbf{58.78} & 64.62 & 28.04 & 39.93\\
        
       & Low & 57.71 & \textbf{58.88} & 58.29 & \textbf{68.61} & \textbf{36.60} & \textbf{45.64}\\
       \midrule    
       GPT-4 &  High & \textbf{30.00} &  \textbf{34.46} & \textbf{32.08} & 2.53 & \textbf{5.90} & 26.37\\
        
       & Low & 29.13  & 34.01 & 31.38 & \textbf{2.72}  & 4.04 & \textbf{29.02}\\
        \bottomrule
    \end{tabular}}
    \caption{
    Performance of various models on Task 1 and Task 2 at different levels of writing quality.}
    \label{table:Impact of Writing Performance}
\end{table}

\subsection{The Relationship between Argumentative Features and Discourse Relations}

\begin{figure}[t]
\centering
\includegraphics[width=0.9\columnwidth]{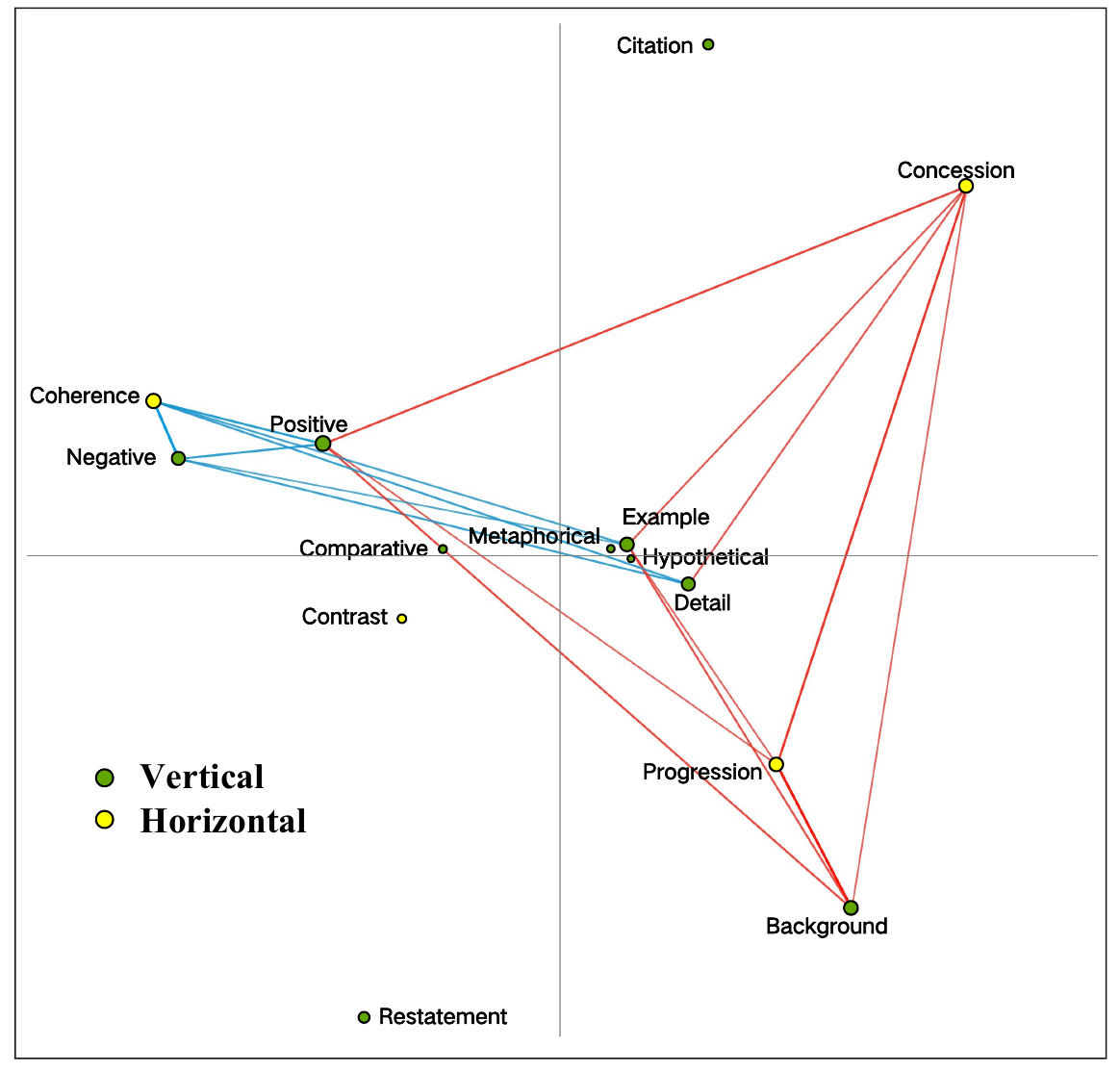} 
\caption{ENA networks of discourse and argument relations in high- (\textcolor{red}{red}) and low-quality essays (\textcolor{cyan!50}{blue}).}
\label{fig: relarions ena}
\end{figure}

As shown in Figure \ref{fig: relarions ena}, the ENA result reveals significant differences in the use of discourse and argument relations between high- and low-quality essays. \textbf{High-quality essays} are more likely to use \textit{Concession} and \textit{Progression} discourse relations, closely integrating \textit{Positive}, \textit{Example}, \textit{Detail}, and \textit{Background}, which presents a logically progressive argumentation. This suggests that high-quality prefer to directly support claims using positive examples, progressively developing the argument through background information and detailed elaboration, while employing concessive relations to enhance depth and critical reasoning. In contrast, \textbf{low-quality essays} primarily focus on \textit{Coherence}, closely combining \textit{Positive}, \textit{Negative}, \textit{Example}, and \textit{Detail}. This suggests that low-quality essays rely on basic parallel reasoning, such as balancing positive and negative argumentation, and providing support through examples and details. 

Overall, high-quality essays demonstrate more critical and hierarchical use of discourse relations, incorporating rich argument relations to effectively enhance the persuasiveness of the reasoning. Conversely, low-quality essays tend to rely on simple and straightforward parallel logic, limiting the depth and effectiveness of argumentation. This finding further validates the result in Section \ref{dis 1}, namely that the complex and diverse relations and structures in high-quality argumentative writing pose greater challenges to the analytical capabilities of models. Therefore, future research could enhance model performance in argumentation analysis by integrating writing proficiency with fine-grained argumentative features, thereby providing more interpretative and comprehensive support for intelligent writing education.

\section{Conclusion}

In this paper, we propose an innovative relation annotation scheme
to characterize the argument strategies and patterns within the CEAMC \cite{ren2024ceamc}. It integrates argument and discourse relations, covering 14 fine-grained relation types from both vertical and horizontal dimensions, thereby overcoming the simplicity and monotony of argument relations in previous studies. We conducted experimental analyses on three tasks, and the results revealed significant differences between PLMs and LLMs across different tasks, indicating that specific tasks require comprehensive consideration of factors such as domain-specific, task complexity, data scale, and model capabilities. Furthermore, additional discussions highlight the importance of fine-grained annotations for a comprehensive understanding of arguments, emphasizing the need for multi-dimensional argument analysis.

\section*{Limitations}

The limitations of our corpus include:

\begin{itemize}
\item \textbf{Data Scale} While our dataset already contains a comprehensive representation of types, it remains limited in size. The diversity and complexity of argumentation imply that the larger the dataset, the more comprehensive its coverage of these phenomena. Consequently, the current size of our dataset might limit the performance and generalization of models trained on it.

\item \textbf{Manual Annotation} Our dataset relies significantly on manual annotations by linguistic experts. Nonetheless, due to the labor-intensive and time-consuming nature of this process, there are inevitable limitations on the volume of annotated data. Further, the inherent subjectivity of manual annotation might lead to potential inconsistencies and bias in the annotated labels.

\end{itemize}


\bibliography{main}

\appendix

\section*{Appendix}
\section{More Details of Annotation Scheme} \label{Annotation Scheme}

By deeply integrating argument relations with discourse relations, we propose 14 fine-grained relation types from both vertical and horizontal dimensions, thereby capturing the intricate interplay between argument components for a thorough understanding of argument structure. Detailed definitions and examples of argument relations in the vertical dimension are presented in Table \ref{table:Argument relation descriptions and samples} and discourse relations in the horizontal dimension are presented in Table \ref{table:discourse relation descriptions and samples}.


\section{More Details of Essay Grading} \label{Details of Essay Grading}

The source corpus CEAMC \cite{ren2024ceamc} consists of 226 argumentative essays from high school examination context and includes essay scores. Table \ref{table:score distribution} details the score distribution, with score ranges categorized according to authoritative scoring criteria. In Section \ref{task}, for the task of Automated Essay Grading, we adopted the original Writing Score Levels, classifying scores into 4 categories. In Section \ref{Discussion} , during the discussion and analysis, due to the limited amount of data and the imbalanced distribution of categories, we combined Category I and Category II essays into a high-quality group, and grouped Category III and Category IV essays into a low-quality group to provide insights into the refined argumentative information at both the high-quality and low-quality levels.

\begin{table}[hbt]
    \centering
    \resizebox{\linewidth}{!}{
        \begin{tabular}{lccc}
            \toprule
            \textbf{Writing Score Level} & \textbf{Counts} & \textbf{Writing Quality Group} & \textbf{\% of Total} \\
            \midrule
            I (63 - 70) & 21 & \multirow{2}{*}{High-quality} &  \multirow{2}{*} {34.51\%} \\
            
            II (52 - 62) & 57 &  & \\
           \midrule
            III (39 - 51) & 146 & \multirow{3}{*}{Low-quality} & \multirow{3}{*}{65.49\%} \\
            IV (21 - 38) & 2 &  &  \\
            V (0 - 20) & 0 &  &  \\
            \midrule
            Total & 226 & - & 100.00\% \\
            \bottomrule
        \end{tabular}
            }
         \caption{Distribution of argumentative essay scores in the research data.}
    \label{table:score distribution}
\end{table}

\section{Concept of ENA} \label{ENA}

Epistemic Network Analysis (ENA) describes the co-occurrence structure within discourse data or any paragraph-based textual data \cite{shaffer2016tutorial}. It has become a mainstream research method in the field of learning analytics, widely used for analyzing and modeling the relationship within collaboration, learning, and cognitive activities \cite{elmoazen2022systematic}. ENA helps researchers understand complex cognitive and interactive processes by visualizing the co-occurrence relationships among different elements.

In ENA, nodes represent distinct concepts, behaviors, or themes. In the context of this study, nodes represent specific \textbf{argument relations} or \textbf{discourse relations}. The edges connecting these nodes indicate the co-occurrence relationships between them, with the strength of these connections represented by the weight (or thickness) of the edges. The edge weight reflects the frequency of co-occurrence between two nodes, where a greater weight indicates a stronger relationship.

\begin{table*}[htb]
\centering
\small
\resizebox{\linewidth}{!}{
\begin{tabular}{l|p{0.1\textwidth}|p{0.3\textwidth}|p{0.4\textwidth}}
\toprule
\textbf{Aspect} & \textbf{Label} & \textbf{Definition} & \textbf{Example}  \\
\midrule
Stance-Based  & Positive       & A method that directly validates the correctness of a viewpoint by using elaboration or evidence consistent with the viewpoint to support it, emphasizing direct affirmation of the viewpoint. \newline   & same with  Metaphorical 

\\

                    & Negative           & A method that indirectly proves the correctness of a viewpoint through elaboration or evidence that are contrary to the viewpoint. It emphasizes the negation of opposing viewpoints, thereby achieving the purpose of the argumentation.   \newline  & \textcolor{blue}{Quotation}: As Shakespeare said, "Without surprises, life would have no luster." \textcolor{red}{->}

\textcolor{blue}{Claim}: Under a certain sense of ceremony, people can become more passionate about life, helping them cherish the moment and look forward to the future.       \\
                      & Comparative        & A shorthand for positive and negative argumentation, is an argumentative approach that involves contrasting and comparing two items to highlight their differences, thereby making the conclusion more evident and persuasive.      &  \textcolor{blue}{Fact}:Take the recent marathon as an example: many contestants did not finish the race, some even quitting midway. This occurred because one runner started accelerating early on, prompting others not to fall behind, a manifestation of tension. Conversely, those who maintained their composure and were undisturbed ended up securing better positions, illustrating the benefits brought by a sense of relaxation.  \textcolor{red}{->}     
                      
                     \textcolor{blue}{claim}: In real life, we need a sense of relaxation more than tension. \\
\midrule
       Evidence-Based                & Example            & An argumentation method that proves a thesis through concrete, or typical examples. \newline 
      & same with  Comparative       \\
                           & Citation           & An argumentation method that proves a thesis by using quotations or axioms.
      & same with  Negative       \\
\midrule
    Discourse-Based        & Metaphorical       &  By employing metaphorical rhetoric, familiar things are used as metaphors to argue the correctness of a viewpoint. In drawing parallels between two items with similar characteristics, the artful use of metaphors often serves to better elucidate concepts, making the argument more vivid and interesting.  \newline & \textcolor{blue}{Elaboration}: If understanding objects is likened to baking a cake, then the method of comprehension is the mold. Those who only heed the words of authoritative experts apply others' molds; thus, no matter how sweet the resulting cake is, it will not be in a shape that suits them. \textcolor{red}{->} 

\textcolor{blue}{Claim}: A deep-rooted reliance on authoritative experts also reflects a more profound issue – a lack of fundamental methods for understanding things oneself.

    \\
                         & Hypothetical       &  Analyzing evidence from the opposite side based on hypothesis to infer its authenticity and reliability, thus robustly supporting a thesis.  \newline     & By employing metaphorical rhetoric, familiar things are used as metaphors to argue the correctness of a viewpoint. In drawing parallels between two items with similar characteristics, the artful use of metaphors often serves to better elucidate concepts, making the argument more vivid and interesting.     
                         
                         \\
                             & Restatement        & For argument of the type \textit{restated claim}, its relation with the target argument (\textit{major claim} or \textit{claim}) is defined as restatement relation.  \newline     & \textcolor{blue}{Restated Claim}: Rituals are never unnecessary or superfluous. \textcolor{red}{->}

\textcolor{blue}{Major claim}: In life, rituals are just so indispensable.     

\\
                             & Detail             & When an argument (\textit{elaboration} type) primarily aims to further explain or analyze other content, it establishes a detail relation with the corresponding argument (\textit{assertion} or \textit{evidence} type).   \newline   & \textcolor{blue}{Elaboration}: Nietzsche's words actually tell us to know thyself and become thyself, which all but maps out the exploration of the spiritual world of self. \textcolor{red}{->}
                             
                             \textcolor{blue}{Quotation}: Nietzsche once said, “Every day that you don't dance is a failure of life.
                             
       \\
                              & Background         &  When an argument (\textit{elaboration} type) primarily serves the function of introducing background, it constructs a background relation with the corresponding argument (\textit{assertion} or \textit{evidence} type).       & \textcolor{blue}{Elaboration}: It's just that is such a mode of exploration really beneficial to people's perceptions?”  \textcolor{red}{->}

\textcolor{blue}{Claim}: This process of transformation essentially reflects the expansion of instrumental rationality and people's active abandonment of “thinking”.
        \\

\bottomrule
\end{tabular}
}
\caption{A list of argument relations in the vertical dimension, their descriptions and samples. Argument component types are indicated in \textcolor{blue}{blue}, with the argument before and after the \textcolor{red}{->} corresponding to the source argument and target argument, respectively. It is noteworthy that multiple argument relations may exist between argument-pair and occurs between argument components of the different type.}
\label{table:Argument relation descriptions and samples}
\end{table*}

\begin{table*}[htb]
\centering
\small
\resizebox{\linewidth}{!}{
\begin{tabular}{l|p{0.3\textwidth}|p{0.4\textwidth}}
\toprule
\textbf{Label} & \textbf{Definition} & \textbf{Example}  \\

\midrule
 Coherence          & Describing several aspects of the same event, related events, or contrasting situations that coexist, co-occur, or oppose in meaning. These aspects can be reordered without altering the overall significance.    \newline    & \textcolor{blue}{Fact}: The idea of a commonwealth of nations, as proposed by Confucius, is also what we aspire to nowadays. \textcolor{red}{->}

\textcolor{blue}{Fact}: Another example is Wang Mang's seizure of power and his promulgation of a series of new measures, which were denied at the time, but in fact he referred to Western countries for these initiatives.

\\
                            Progression        & The subsequent argument represents an advance in scope or meaning than the preceding one, intended to emphasize a deepening, expansion, or reinforcement of logic, and the order of the arguments is usually non-interchangeable.  \newline  
      & \textcolor{blue}{Claim}: However, the negative impacts caused by the pursuit of rituals are not few.  \textcolor{red}{->}

\textcolor{blue}{Claim}: Only by getting rid of the solidified idea that a sense of ritual is necessary in life can they focus on the abundance of the spiritual world and climb higher.       \\
                          Contrast           & Comparison and selection are made by examining the similarities or differences between two or more things, situations, or viewpoints, emphasizing the contrast between them.   \newline      & \textcolor{blue}{Fact}: We all know that Wei Liangfu improved the Kunqu opera, leaving brilliant cultural treasures for future generations, we all know that Yuan Longping broke through a technical barrier to solve the food problem in many areas, they are not precisely in the ancients and the authority of the forefathers under the influence of their own chapter? \textcolor{red}{->}

\textcolor{blue}{Fact}: There are great men, naturally, there are also small people, those so-called good learning in fact, “thick ancient and thin” academic molecules, those who listen to the authority of the scientific molecules do not understand the development of adaptability, which one has made achievements?    

\\
                        Concession         & An argument posits a certain situation or viewpoint, followed by a shift where the subsequent argument presents an opposing or contrasting perspective, emphasizing the content of the latter argument.     &  \textcolor{blue}{Claim}: Therefore, while inheritance is important, breakthroughs and development are also indispensable. \textcolor{red}{->}

 \textcolor{blue}{Claim}: However, should those ideas and factors that have been tested be recognized in their entirety? No.

 \\
\bottomrule
\end{tabular}
}
\caption{A list of discourse relations in the horizontal dimension, their descriptions and samples. Argument component types are indicated in \textcolor{blue}{blue}, with the argument before and after the \textcolor{red}{->} corresponding solely to the order in which the two arguments appear in the essay. It is noteworthy that the discourse relations between argument-pair is singular and occurs between argument components of the same type.}
\label{table:discourse relation descriptions and samples}
\end{table*}

\end{document}